# Toward Real-Time Decentralized Reinforcement Learning using Finite Support Basis Functions

Kenzo Lobos-Tsunekawa, David L. Leottau, and Javier Ruiz-del-Solar

Advanced Mining Technology Center & Dept. of Elect. Eng., Universidad de Chile
{kenzo.lobos,dleottau,jruizd}@ing.uchile.cl

**Abstract.** This paper addresses the design and implementation of complex Reinforcement Learning (RL) behaviors where multi-dimensional action spaces are involved, as well as the need to execute the behaviors in real-time using robotic platforms with limited computational resources and training times. For this purpose, we propose the use of decentralized RL, in combination with finite support basis functions as alternatives to Gaussian RBF, in order to alleviate the effects of the curse of dimensionality on the action and state spaces respectively, and to reduce the computation time. As testbed, a RL based controller for the in-walk kick in NAO robots, a challenging and critical problem for soccer robotics, is used. The reported experiments show empirically that our solution saves up to 99.94% of execution time and 98.82% of memory consumption during execution, without diminishing performance compared to classical approaches.

## 1  Introduction

Reinforcement Learning (RL) has been applied in robotics for learning complex behaviors. However, even though successful implementation of RL in robotics have increased largely in the last few years [18], there are still many factors that limit the massive use of RL in general robotic problems, such as the high dimensionality of the state and action spaces, and the need of numerous real-world experiments.

The first factor is related to the inherent complexity of behaviors required by robots operating in the real world. Thus, when fine-grained RL controllers are required to achieve high performance, classic RL models scale exponentially their complexity, according to the dimensionality of the state and action spaces. This is undesirable since it causes two main effects: an unfeasible number of training episodes to achieve asymptotic convergence, and an impracticable computational cost for real-time execution, since the controller actions usually need to be executed in short time cycles to achieve high performance. In light of these issues, the RL approach becomes limited to those problems in which high computation resources are available, long experiment and training times are possible, and/or the design or modeling can be simplified without sacrificing performance.

In this work, we propose to deal with these issues, without oversimplifying the problem itself, by: (i) using Decentralized Reinforcement Learning (D-RL), which empirically accelerates convergence [4] and reduces considerably the effects of the curse of dimensionality on the action space; and (ii) applying Finite Support Basis Functions (FSBF) in the state representation of RL, which provides a representation similar as the one resulting from the use of Gaussian RBF (Radial Basis Functions), but with much reduced computation time.

To validate our approach, we consider the specific case of the in-walk kicks [12] in robotic soccer, where the robot must learn to push the ball toward a desired target, only by using the inertia of its own gait. This problem becomes an ideal testbed, since it is a challenging and relevant problem in robotic soccer, and both the state and action spaces must be sufficiently fine-grained to capture the complexity of the problem and achieve competitive performances. Moreover, the solution must run on real time, on an embedded device such as the NAO robot's CPU [19].

The proposed model for the in-walk kick problem is similar to the one presented in [9] and [6], since this task can be considered as a particular case of the ball-pushing behavior. However, those works differ from ours, since for this case, all three axes of the gait velocity vector are learned autonomous and simultaneously through RL, and no human intervention is required during the learning process. In addition, we introduce a novel approach to adapt and synchronize the RL time step to the time step of the robot's gait.

The remainder of this paper is organized as follows: D-RL is reviewed shortly in Section 2. Section 3 presents a brief overview about FSBF for the proposed state representation. Then, the in-walk kick behavior and the proposed modeling are described in Section 4. Finally, Section 5 presents an experimental validation of the proposed methodology, and in Section 6 conclusions of this work are given.

## 2  Decentralized Reinforcement Learning

Two main limitations were previously remarked about the use of RL in complex behaviors performed with physical robots: the high number of training episodes to achieve asymptotic convergence, and the expensive computational cost for real-time execution. Under the presence of multi-dimensional action spaces, the standard RL solutions can be called Centralized RL (C-RL) systems, if each of those action subspaces are discretized (in a finite set of actions), combined, and computed as a single set of actions.

In C-RL, the number of possible actions grows exponentially with the action space dimensionality. This makes it hard exploring sufficiently the whole action-state space, producing a very slow convergence, and increasing exponentially the execution time. D-RL helps to alleviate both issues, by splitting the learning problem into multiple independent agents, each acting in a different action space dimension [4]. This allows the design of independent state models, reward functions, and learning agents for each action dimension. The benefits of the D-RL over C-RL in terms of computation time can be quantified as [2]:

$$DRL_{speedup} = \frac{\prod_{m=1}^{M} A^m}{\sum_{m=1}^{M} A^m} \qquad (1)$$

where $M$ is the number of agents (action space dimensionality), and $A^m$ the number of possible discrete actions in the action dimension $m$.

In [2], a robotic application of D-RL is presented, where the individual action components are learned in parallel by independent learning agents. Moreover, [4] shows empirically that D-RL is not only capable of attain coordination among agents without any explicit mechanism of coordination, but that D-RL outperforms C-RL in two different problems: a modified version of the popular MountainCar3D testbed, and a physic robot tasked with a ball-pushing problem.

## 3    Finite Support Basis Functions

One important design component of a RL agent is how it deals with the state space. While for discrete state space problems the classic tabular representation is natural, other representations are preferred for continuous, or discrete but large, state spaces.

For real-time applications, linear parametric approximations are usually used, being among the most used tile coding [16] and Gaussian RBF [14]. Although Gaussian RBF have many desired mathematic properties, their use is usually restricted due to its high computation time, and therefore the fast tile coding is used, even when Gaussian RBF can achieve a better performance [15].

To deal with this issue we propose the use of FSBF, as replacements of Gaussian RBF to provide a similar state space representation, with a much-reduced computation time, due to the sparsity introduced by these FSBF in the feature vector. Although the approximation of Gaussian RBF has been addressed in [7, 17], mainly in the field of control systems, the computational benefits of this approximation have not been presented, and no previous applications of this kind of approximation have been presented in the context of RL, where the speedup produced by its use can be very large, due to the need to compute the Q-Values for every possible action.

Consider the case of a Gaussian RBF representation, in which Gaussians are placed in a uniform fashion over the state space $S$, with dimensionality $|S| = N$. In each dimension $i$, $n_i$ 1D Gaussians are used, creating a total amount of $M = \prod_{i=1}^{N} n_i$ multivariate Gaussians. Then, the Q-Values needed for value iteration algorithms are computed as:

$$Q(s,a) = \frac{\sum_{j=1}^{M} \phi_i(s) \theta_j^a}{\sum_{j=1}^{M} \phi_i(s)} \qquad (2)$$

where $\theta^a$ are the weights for action $a$, and $\phi$ is the $M$-dimensional feature vector.

The main issue with linear Gaussian RBF approximators, is that $M$ grows exponentially with $N$, and since Gaussians are non-zero over the entire space, all the elements in the sum must be calculated explicitly, even if only a few of them actually contribute to the sum. However, if multivariate Gaussians are chosen with the same diagonal covariance matrix, equation (2) can be rewritten as:

$$Q(s,a) = \frac{\sum_{k \in A}\left(\prod_{n=1}^{N} \phi^i_{k_i}(s_i)\right)\theta^a_k}{\sum_{k \in A}\left(\prod_{n=1}^{N} \phi^i_{k_i}(s_i)\right)} \qquad (3)$$

where $\phi^i_{k_i}$ are 1D Gaussians, and $A = \{(k_1, \ldots, k_N) | \ k_i = 1, \ldots, n_i \ \ i = 1, \ldots, N\}$.

From equation (3), it can be noted that the number of different 1D Gaussians $\sum_{i=1}^{N} n_i$ is much lower that the number the $\prod_{i=1}^{N} n_i$ multivariate Gaussians, and that the same 1D basis function appears many times in the sum.

It follows that if the unidimensional basis is chosen with finite support region, then the number of non-zero terms in the sum becomes strongly reduced, thus saving much computation time. Some examples of FSBF are 3-$\sigma$ Gaussians approximations, and kernels like Epanechnikov, cosine and even triangular functions, as shown in Figure 1.

Using this methodology, the computation cost depends only on the width chosen for the unidimensional basis functions, and their density in the state space, instead of the number of basis functions. This allows to place more basis functions over the state space without extra computation time, if their density and support region width are kept. Finally, the computation speedup from using this strategy can be expressed as:

$$state_{speedup} = \frac{\prod_{i=1}^{N} n_i}{\prod_{i=1}^{N} width_i} \qquad (4)$$

where $width_i$ is the support region width of the basis functions in the dimension $i$, in terms of the distance among basis functions in the same dimension.

In Section 5, the proposed state representation is empirically validated with the in-walk kick problem by comparing it with the standard Gaussian RBF representation.

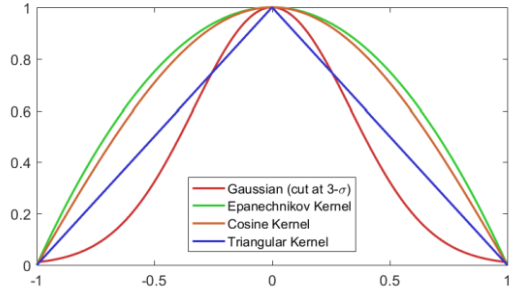

*Figure 1: Normalized finite support functions.*

## 4    Validation Problem: The In-Walk Kick in Biped Robots

In the context of robotic soccer, kick motions can be considered the second most important robot motion after walking, due to their impact in the game performance. Given that, several methods have been developed for the design and implementation of kick motions [1,10,12,13]. These methods can be classified in two groups: methods that create independent motions [10,13,1], and methods that make use or modify the robot gait to perform a kick (in-walk kicks) [11]. While the former methods are able to make full use of the hardware capabilities of the robot and can be studied inde-pendently, the later have the advantage of a much faster activation time, since they do not need to wait for the robot to finish the walking motion and to stabilize, saving critical time in challenging game situations. The in-walk kicks were originally introduced and implemented in [12], in which the gait phases are modified to create kick motions. To the best of our knowledge no other implementations have been reported. It must be noted that the in-walk kicks approach is not only limited by the hardware capabilities of the robot, but also by the gait design and implementation.

In this work, we propose a variant of the in-walk kick [12], which is commonly used in the Standard Platform League (SPL) of the RoboCup competition[1]. The proposed implementation consists of in-walk kicks that are performed using only the inertia of the gait, without any specially designed kick motion. Details of the modeling and implementation are presented below.

### 4.1 General Modeling

We propose an end-to-end implementation of the in-walk kicks, by designing a RL based controller over the omnidirectional gait developed in [5] as released in [11], i.e. designing a controller over the velocities commands $v_x$, $v_y$ and $v_\theta$.

The proposed in-walk implementation can be viewed as an instance of the ball-pushing behaviors proposed in [6] and [9], and the same state and action models can be used. These models have been thoroughly validated in many implementations of the ball-dribbling behavior (another example of ball-pushing behavior), such as 2D simulators [6,8], 3D simulators [9,3,7] and on physical robots [9,3], obtaining excellent results in the RoboCup 2015 and 2016, in the SPL league.

The basic state model of the ball-pushing behaviors presented in [6] consists on the distance from the robot to the ball ($\rho$), the bearing of the robot with respect to the ball ($\gamma$), the angle formed between the robot, the ball and the target ($\varphi$), and the distance from the ball to the desired target ($\psi$). All these distances and angles are calculated from a foot coordinate frame, and the specific foot is chosen by a handcrafted foot selector. The angles and distances are shown in Figure 2.

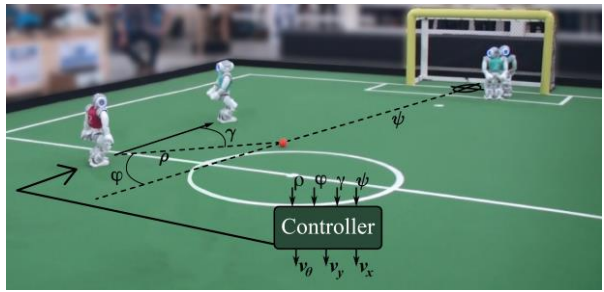

*Figure 2: Geometric state variables for the NAO using a magenta jersey. Obstacles, such as the NAO robots using cyan jerseys are not considered in the state formulation.*

---

[1] RoboCup SPL official web site: http://www.tzi.de/spl/bin/view/Website/WebHome

### 4.2 Extended Ball-Pushing Modeling for Humanoid Biped Robots

Although several publications validate that the previous state model can obtain good performances [6,3,8], that modeling does not consider many aspects of the physical robot such as its biped nature and the used gait implementation [5]. We propose the following extensions to the previous model, since in the in-walk kick problem, as opposed to the ball-dribbling problem, the performance is the result of a single interaction between the robot and the ball, thus needing as much precision as possible:

- **Gait Synchronization:** Previous implementations of RL controllers for the ball-dribbling use a fixed time step among actions. However, the implementation of the robotic gait in [5] is constantly planning its next step, and only reads velocity commands once in each phase (half-step). Neglecting this phenomenon derives in taking actions in different parts of the phases, producing a violation of the Markov property. To address this issue, we propose a model in which the controller's actions are taken once per half step, instead of using a fixed time step.
- **Phase Type:** Even if the agent actions are synchronized with the gait generator (walking engine), symmetric situations arise since at the end of each phase, the robot has one foot forward and the other foot backwards, and if this phenomenon is not included in the state model, precise time step planning becomes very difficult. To model this phenomenon, a binary state can be added to the state model to represent which foot is currently forward.
- **Foot Selector:** Even though the foot selector proposed in [9] works well, it can limit the performance, mainly in symmetric situations where the foot selector can arbitrarily assign the farthest foot from the ball. We propose to calculate the features $\rho$, $\gamma$ and $\varphi$ from the center of the robot, instead from the selected foot coordinate system, so the RL agent can learn its own foot selector.

### 4.3 Decentralized Reinforcement Learning Modeling

This work uses the results presented in [8] and [4] which empirically validate the effectiveness of the decentralized approach in mobile robotics and common testbeds.

The design of the action space while critical, maintains the approach presented in [8]. Since the requested velocity vector of the omnidirectional biped walk engine is $[v_x, v_y, v_\theta]$, it is possible to decentralize this 3-Dimensional action space by using three separate learning agents, $Agent_x$, $Agent_y$, and $Agent_\theta$ acting in parallel in a multi-agent task. Careful considerations must be taken in the number of discrete actions for each agent. While larger numbers of actions produce more complex behaviors, it also increases the number of episodes needed for convergence, since not only the search space grows, but also coordination among agents is more difficult. The proposed action space is shown in Table 1, where the minimum and maximum values for each velocity component are the ones used in [11].

The proposed state model shared by the three independent agents is presented in Table 1, where the geometric state variables $\rho_{center}$, $\gamma_{center}$ and $\varphi_{center}$ are calculated from the center of the robot, $phaseType$ represents which foot is currently forward, and the number of cores corresponds to the number of 1D basis functions placed uniform along each state dimension (the product of the number of cores becomes the dimensionality of the feature vector).

*Table 1. State and action spaces for the D-RL modeling of the in-walk kick problem.*

| Joint state space: $S = [\rho, \gamma, \varphi]^T$ | | | | Action space: $A = [v_x, v_y, v_\theta]^T$ | | | |
|---|---|---|---|---|---|---|---|
| Feature | Min | Max | N.Cores | Agent | Min | Max | N.Actions |
| $\rho_{center}$ | 0mm | 800mm | 15 | $v_x$ | 0mm/s | 120mm/s | 16 |
| $\gamma_{center}$ | -70° | 70° | 11 | $v_y$ | -70mm/s | 70mm/s | 15 |
| $\varphi_{center}$ | -90° | 90° | 13 | $v_\theta$ | -30°/s | 30°/s | 17 |
| $phaseType$ | - | - | Binary | | | | |

Finally, the reward is chosen the same for each independent agent, to ensure a full-cooperation task. Since the objective is to kick as strong and accurate as possible, a high positive reward must be given at the terminal state when the robot hits the ball, but decay depending on the ball trajectory deviation from the target, and the distance between the ball and the target. For non-terminal states, a reward similar to the ball-dribbling problem [6] is used to guide the robot towards the ball:

$$r = \begin{cases} K \exp\left(-\frac{\psi_{error}}{\psi_0}\right) \exp\left(-\frac{\alpha_{error}}{\alpha_0}\right) & \text{if ball touched} \\ -\left(\frac{\rho}{\rho_{max}} + \frac{|\varphi|}{\varphi_{max}} + \frac{|\gamma|}{\gamma_{max}}\right) & \text{otherwise} \end{cases} \quad (5)$$

where $\psi_{error}$ is the distance that the ball still needs to travel to reach the target in its current trajectory, $\alpha_{error}$ is the angle deviation of the ball's trajectory from a straight line to the target, and the parameters $\psi_0$ and $\alpha_0$ allow the design of rewards with more focus on kick strength or precision. Finally, the gain parameter $K$ is used for convergence purposes, since the high positive reward can only be given once per episode.

## 5 Results

### 5.1 Experimental Set Up

We present experiments that compare the proposed state model presented in Section 4 (gait synchronization, phase type and no foot selector) against the one presented in [8], as well as the use of the different basis functions presented in Section 3. It is worth mentioning that in these experiments we use just the proposed D-RL modeling and not C-RL, since in the in-walk kick with biped robots problem the action space is large enough to make the use of C-RL unfeasible (with more than a thousand discrete actions).

The experiments are performed in the SimRobot 3D simulator released in [11]. Since the objective of the training is to achieve fast yet powerful and accurate kicks, with the intent to score goals in as many situations as possible, the training conditions must be both challenging and general. In each episode, the ball is situated on a fixed position from the goal, with a distance of 1.5m, which is the experimental maximum distance that a linear controller can achieve in the current environment. Furthermore, the robot starts in a random position, at 1m from the ball and always facing it, but with a $|\varphi| < \varphi_{max}$, as shown in Figure 3. This random initialization is designed so the learning agent can successfully learn many points of operation, and thus achieve a general kick behavior. The episode termination criterion is given by the following conditions: episode timeout of 200 seconds, the robot or the ball leaves the field, or the ball is pushed by the robot. Since episodes can finish in less than a few seconds and with unsuccessful kicks, a high number of episodes is needed. In the following experiments, the total number of training episodes is set to 1,500.

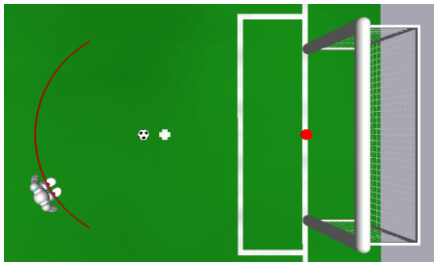

*Figure 3. Training environment. The red segment represents possible initial positions of the robot, and the red circle is the kick target.*

*Table 2. Training parameters.*

| Parameter | Value |
|---|---|
| Learning rate | 0.1 |
| Discount factor | 0.99 |
| $\lambda$ | 0.9 |
| $\sigma$ | $0.5\, \Delta$ |
| $width$ | $3\, \Delta$ |
| $decay$ | 15 |
| $K$ | 50 |
| $\psi_0$ | 300mm |
| $\alpha_0$ | 14° |

The performance indexes considered for the experiments are defined as follows:
- **DistanceError** ($\psi$)**:** The distance left to travel from the ball to the goal line, normalized by the initial 1.5m distance. This measures the ability to perform strong kicks.
- **AngleError** ($\alpha$)**:** The angle deviation of the ball's trajectory from a straight line to the target, normalized by the maximum angle that would score a goal. This measures the accuracy of the kicks, and depends on the previous alignment.

The following experiments are performed using the popular $Sarsa(\lambda)$ algorithm for training each of the three agents, with an exponentially decaying $\epsilon$-greedy exploration ($\epsilon = \exp(-decay * episode/maxEpisodes)$). Since the idea of using FSBF to save computation time needs to be validated, experiments are performed using the following functions as 1D basis functions: cosine kernel, Epanechnikov kernel, triangular kernel, and 3-$\sigma$ Gaussian approximation. The parameters used in the experiments are shown in Table 2, where $\Delta$ is the distance between the centers of the basis functions in each dimension, and the parameters $K$, $\psi_0$ and $\alpha_0$ correspond to the parameters of the reward in equation (5). The value of $\sigma$ is a very common heuristic choice, and the support area of the other FSBF is set to match the 3-$\sigma$ Gaussian approximation in terms of support area, to provide comparative results against fixed computation resources. For illustrative purposes, a video showing some episodes of the learning evolution is shown in [20], in which the proposed state model and the Epanechnikov basis functions are used.

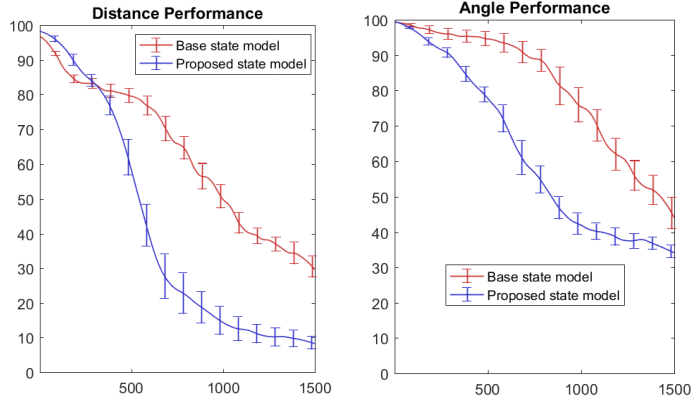

*Figure 4. Training performances for the different state models.*

## 5.2 State Model Comparison

The experiments in Figure 4 show the learning evolution using the base state model for the ball-pushing behaviors [8] and the one proposed here. In both cases, the state representation is done using Gaussian RBF, and the rest of the parameters are taken from Table 2. The learning evolution results are plotted by averaging 15 trials, and error bars show the standard error.

These results show a significant improvement when using the proposed model. The reason for the difference in the distance performance is explained by the fact that in the base model a precise step planning could not be performed, and thereby, the kicks lose strength. The differences in the angle performances are explained by the lack of foot selector and the inclusion of the *phaseType* in the proposed state model, which allows the robot to align itself correctly to kick the ball, even in symmetric conditions, where the previous foot selector choses arbitrarily foot assignments.

## 5.3 Basis Functions Comparison

To provide results for the proposed state representations, Figure 5 presents the learning evolution of different finite basis functions, as well as the traditional Gaussian RBF representation. In the presented cases, the proposed state model is used, and the rest of the parameters were taken from Table 2. Again, the learning evolution results are plotted by averaging 15 trials, and error bars show the standard error.

The plots in Figure 5 show that the FSBF are a competitive choice against Gaussian RBF representations. While the 3-$\sigma$ Gaussian approximation maintains the performance of standard RBF representation, the other basis functions, specially cosine and Epanechnikov clearly outperform the Gaussian representation.

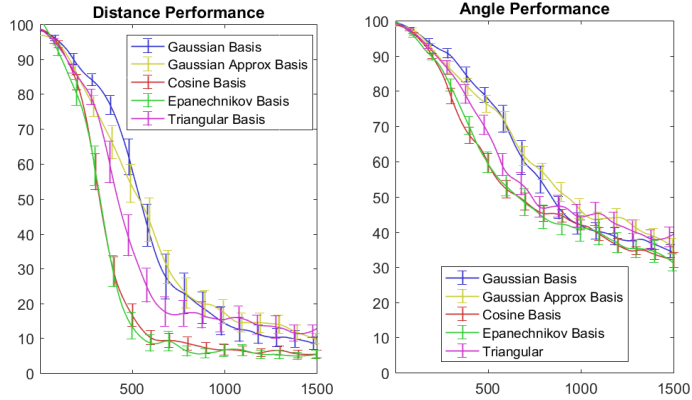

*Figure 5. Training performances for the different basis functions.*

*Table 3. Execution times for the different approaches.*

| Setting | Execution Time [ms] | Model Size [MB] |
|---|---|---|
| C-RL + Gaussian RBF | 304.72 | 70 |
| C-RL + FSBF | 11.57 | 70 |
| D-RL + Gaussian RBF | 4.42 | 0.82 |
| D-RL + FSBF | 0.17 | 0.82 |

### 5.4 Execution Time and Memory Consumption

Since the proposed methodology must run in real time on an embedded system, its execution time must be negligibly compared to the 17ms cycles produced by the image acquisition rate in the NAO, and the model must be small enough to fit in the available memory. Table 3 presents the execution times and memory consumptions of different approaches. This table also considers C-RL to remark the importance of the proposed methodology in reducing computational resources. Memory consumption is measured from model's weights, and execution time is measured as the average execution time of a random policy, since the execution time does not depend on the learned policy, and using random policies allows the inclusion of the C-RL approach in the analysis, even if it was not trained. The FSBF used in Table 3 corresponds to the 3-$\sigma$ Gaussian approximation, and the executions times for other FSBF are very similar, since the evaluation time of the 1D FSBF is negligibly compared to the feature vector calculation. To provide comparable results, in all cases the number of discrete actions per action dimension and cores are kept the same.

The results in Table 3 show that substantial computation resources can be saved by using the proposed methodology. The effective speedup produced by the D-RL against its C-RL counterpart comes close to the theoretical one expressed in Equation (4) (an effective speed of x69 against the theoretical of x85), and the effective speedup from the state representation is x26 instead of x238, since memory access for FSBF is not contiguous, among other overheads. Finally, a x85 memory saving is achieved product of the D-RL scheme, with models using less than a megabyte of memory, fitting in most embedded systems.

## 6  Conclusions

In this paper, we presented a general methodology for achieving high performance and computationally cheap RL behaviors, for complex problems in which both, training times and computation resources are limited. In that sense, we presented the following solutions: first, to alleviate the effects of the curse of dimensionality on the action space, Decentralized RL was proposed, where each action dimension is controlled individually by independent learning agents acting in parallel in a multi-agent task. Second, the use of finite support basis functions, as alternatives to Gaussian RBF as linear approximators, were proposed to generalize the state space and reduce considerably computation time.

We considered the in-walk kick task from soccer robotics as a real-world validation problem. The proposed solution saves about 99.94% of execution time and 98.82% of memory consumption with respect to a centralized RL system implemented with standard Gaussian RBF, without sacrificing performance. Furthermore, basis functions like cosine and Epanechnikov empirically showed better performance and faster convergences than the Gaussian RBF representation.

We also presented a novel state model for the ball-pushing behaviors in which the RL agent actions are synchronized with the biped gait. This outperformed considerably the in-walk kick in terms of accuracy and effectiveness with respect to the former modeling for ball-pushing behaviors

Finally, it is interesting to mention that the behavior that arises through model-free RL makes intense use of the dynamics and geometry of the robot. In [20] some examples of the learned behaviors can be observed.

## Acknowledgements

We would like to thank the BHuman SPL Team for sharing their code release, contributing the development of the Standard Platform League. This work was partially funded by FONDECYT Project 1161500.